\pdfoutput=1

\documentclass[11pt]{article}

\usepackage[preprint]{arxiv}

\usepackage{times}
\usepackage{latexsym}

\usepackage[T1]{fontenc}

\usepackage[utf8]{inputenc}

\usepackage{microtype}

\usepackage{inconsolata}

\usepackage{graphicx}
\usepackage{amssymb}
\usepackage{epstopdf}
\usepackage{amsmath}
\usepackage{array}
\usepackage{bbding}
\usepackage{colortbl}
\usepackage{subcaption}
\usepackage{caption}
\usepackage{xcolor}
\usepackage{float}

\title{QG-VTC: Question-Guided Visual Token Compression in MLLMs for Efficient VQA}

\author{
    Shuai Li\textsuperscript{\(\S\)}, Jian Xu\thanks{~~~Corresponding Author.}\textsuperscript{\(\diamond\)}, Xiao-Hui Li\textsuperscript{\(\diamond\)}, Chao Deng\textsuperscript{\(\diamond\)}, and Lin-Lin Huang\textsuperscript{\(\S\)} \\
    BeiJing JiaoTong University, Beijing, 100044, China\textsuperscript{\(\S\)} \\
    State Key Laboratory of Multimodal Artificial Intelligence Systems, Institute of \\
    Automation of Chinese Academy of Sciences, Beijing, 100190, China\textsuperscript{\(\diamond\)} \\
    shuai.li@mais.ia.ac.cn, huangll@bjtu.edu.cn, xiaohui.li@nlpr.ia.ac.cn, \\
    \{jian.xu, dengchao2023\}@ia.ac.cn
}

\begin{document}
\maketitle
\begin{abstract}
Recent advances in Multi-modal Large Language Models (MLLMs) have shown significant progress in open-world Visual Question Answering (VQA). However, integrating visual information increases the number of processed tokens, leading to higher GPU memory usage and computational overhead. Images often contain more redundant information than text, and not all visual details are pertinent to specific questions. To address these challenges, we propose \textit{\textbf{QG-VTC}}, a novel question-guided visual token compression method for MLLM-based VQA tasks. \textit{\textbf{QG-VTC}} employs a pretrained text encoder and a learnable feed-forward layer to embed user questions into the vision encoder's feature space then computes correlation scores between the question embeddings and visual tokens. By selecting the most relevant tokens and softly compressing others, \textit{\textbf{QG-VTC}} ensures fine-tuned relevance to user needs. Additionally, a progressive strategy applies this compression across different vision encoder layers, gradually reducing token numbers. This approach maximizes retention of question-relevant information while discarding irrelevant details. Experimental results show that our method achieves performance on par with uncompressed models using just 1/8 of the visual tokens. The code and model will be publicly available on \textit{GitHub}.
\end{abstract}

\section{Introduction}
Large Language Models(LLMs) have demonstrated excellent proficiency in the linguistic modality, exhibiting robust capabilities in understanding,reasoning, and generation~\cite{brown2020language,ouyang2022training,devlin2018bert,ouyang2022training,raffel2020exploring,touvron2023llama,chiang2023vicuna, bai2023qwen}. The advent of Multimodal Large Language Models(MLLMs) have expanded the competencies of LLMs beyond the linguistic modality~\cite{achiam2023gpt,team2023gemini,team2024gemini}, with its effective comprehension of images marking a significant advancement to handle multimodal information. In recent years, numerous high-performance, open-source MLLMs have been introduced consecutively~\cite{li2023blip, ye2024mplug, liu2024visual, wang2024qwen2, liu2024improved}, greatly fostering the development of this field.

Common open-source MLLMs~\cite{liu2024visual,liu2024improved,chen2023llava,liu2025llava,ye2024mplug,wang2024qwen2} typically consist of vision encoder, adapter, and LLM. The vision encoder serves to extract image features, while the adapter is responsible for projecting these image features into semantic space to facilitate the LLM's understanding of image information. Prior to being fed into the LLM, image information is represented as visual tokens. Generally, the higher the resolution of the image, the greater the number of visual tokens, which implies increased memory consumption and computational time during both training and inference. If the total number of visual tokens and textual tokens exceeds the context limit of the LLM, tokens beyond this limit are directly truncated, rendering MLLMs highly susceptible to giving erroneous responses.

Intuitively, compared to the highly condensed linguistic modality of human knowledge, the visual modality often contains more redundant information. Furthermore, when performing question answering tasks based on images, not all visual information is relevant to the specific question at hand. This has prompted some researchers~\cite{chen2023diffrate, shang2024llava, chen2025image, li2025llama, cha2024honeybee, song2024less, chen2024recoverable} to focus on reducing the number of visual tokens in order to enhance the training and inference efficiency in the field of image question answering.

These efforts attempt to compress visual tokens within the LLM~\cite{chen2025image,zhang2024sparsevlm}, between the LLM and the projector~\cite{li2025llama}, between the projector and the vision encoder~\cite{shang2024llava,song2024less,chen2024recoverable}, or by directly designing unique visual projectors~\cite{cha2024honeybee,li2024tokenpacker}. Despite promising results, these methods still face some limitations. Firstly, visual token compression does not consider user queries, leading to indiscriminate data reduction. Ideally, compression should preserve information relevant to the user's question while minimizing less pertinent data. Secondly, compression also occurs too late, missing opportunities to reduce computational load earlier in the vision encoder itself. Furthermore, designing new visual projectors for compression complicates training process and requires more training data.~\cite{yao2024deco}.

To overcome the above limitations, we propose a novel visual token compression method named \textit{\textbf{QG-VTC}} which operates internally and hierarchically within the vision encoder. By computing the similarity between visual tokens and the question text embedding, highly pertinent visual tokens are identified and selected. After that, the other less pertinent visual tokens are softly recycled according to attention scores between them and previous selected tokens. This method effectively reduces the number of visual tokens without incurring excessive loss of image information. Furthermore, it implements hierarchical compression based on the layered structure of ViT, resulting in a smoother and less abrupt compression process. More crucially, it significantly reduces the computational load of the vision encoder itself, which is essential when processing high-resolution images.

The contributions of this paper are summarized as follows:
\begin{itemize}
\item We propose \textit{\textbf{QG-VTC}}, a method for question-guided visual token compression. It selectively retains the most relevant visual tokens based on the user's question and merges less relevant ones using weighted averaging. This approach drastically reduces token count with minimal loss of image information.

\item Our method integrates a question-guided visual token compression module into the vision encoder using a hierarchical strategy. This approach preserves local image details while leveraging deep, semantic-rich information, reducing computational load on both the LLM and vision encoder itself.

\item Through extensive experiments, we achieve SOTA results on multiple benchmarks. Notably, we match performance with only 1/8 the visual tokens and roughly 30\% of the original computational load.
\end{itemize}

\begin{figure*}[htb]
\centering
    \includegraphics[width=0.98\textwidth]{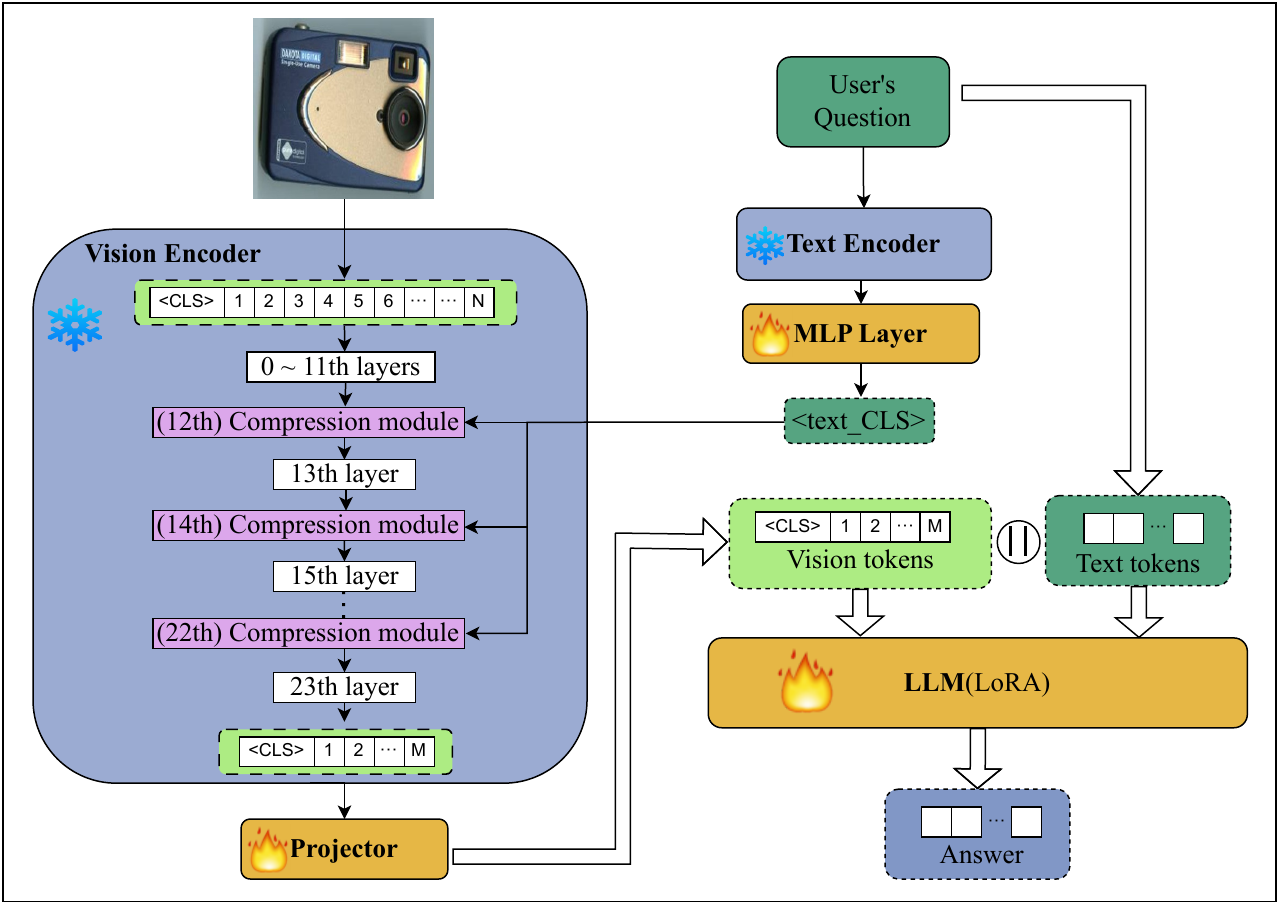}
\caption{The overall architecture of \textit{\textbf{QG-VTC}}. The compression module within the vision encoder is capable of compressing visual information under the guidance of a user's question. The projector is responsible for projecting the compressed visual information into the semantic space of the LLM. Subsequently, the vision tokens and text tokens are concatenated into a sequence, which is then input into the LLM to obtain the answer.}
\label{fig1}
\end{figure*}

\section{Related Work}
\subsection{Multimodal Large Language Models}
Currently, prevalent MLLMs are extensions based on LLMs. To process visual information, MLLMs employ vision encoders to convert images into visual tokens, which are then aligned with the language space through a projector. This alignment enables the LLMs to comprehend the visual content. Early models such as BLIP2~\cite{li2023blip} and InstructBLIP~\cite{dai2023instructblip} employ a frozen ViT-L/14~\cite{dosovitskiy2020image} as vision encoder, and design a Q-Former specifically to bridge the gap between the visual and linguistic modalities. Flamingo~\cite{alayrac2022flamingo} and Qwen-VL series~\cite{bai2023qwenvl, wang2024qwen2} utilize cross-attention modules(alternatively referred to as Resamplers) to extract visual information specifically for their integrated LLMs. These approaches rely on extensive datasets and sophisticated training procedures to achieve effective cross-modal alignment.

CLIP employs contrastive language-image pretraining on a vast dataset of image-text pairs, and provides a series of ViT modules~\cite{radford2021learning} with exceptional cross-modal capabilities. This implies that the visual tokens it generates are closely aligned with the semantic space. Benefiting from this, LLaVA-1.5~\cite{liu2024improved} directly utilizes the ViT module from the CLIP as the vision encoder. It projects the generated visual tokens to the semantic space through a straightforward MLP, and only a limited amount of data is used for pre-training and instruction tuning of the MLP and LLM, with the vision encoder frozen. The thriving development of the LLaVA family~\cite{liu2025llava, guo2025llava, lin2023video, lin2024moe, shi2024math}, coupled with numerous research studies(e.g. \citealp{chen2020simple, chen2020improved, cao2023strip, gao2024transformer}), has demonstrated that combining CLIP-ViT, MLP, and LLMs is a straightforward and effective approach for constructing MLLMs. However, compared to methods such as Resampler~\cite{wang2024qwen2} and Q-former~\cite{li2023blip}, MLP does not compress the number of visual tokens.

\subsection{Visual Tokens Compression}
To alleviate the computational burden associated with the visual modality, reducing the number of visual tokens is an effective approach.

FastV~\cite{chen2025image} analyzes the inefficient visual attention phenomena within LLM and ranks the importance of visual tokens using average attention scores. In the shallow layers of LLM, it reduces the number of visual tokens by half through a pruning approach. PruMerge~\cite{shang2024llava} categorizes visual tokens into unpruned and pruned groups by computing the attention scores between the visual tokens output by the vision encoder and the class token of the image, then cluster the pruned tokens based on key similarity and merge the clustered tokens with the unpruned tokens to supplement their information. Zhang et al.~\cite{zhang2024token} introduce the concept of information density of sub-image, and adaptively compress sub-images into different lengths. Honeybee~\cite{cha2024honeybee} and TokenPacker~\cite{li2024tokenpacker} compress visual tokens by designing novel projectors, which can be viewed as improved versions of methods such as Resampler and Q-former. These works do not utilize user questions as guidance when compressing visual tokens.

Chen et al.~\cite{chen2024recoverable} leverage the similarity between the question text and visual tokens to recover visually meaningful tokens with important text information while merging other less important tokens. FocusLLaVA~\cite{zhu2024focusllava} achieves coarse-to-fine visual token reduction by employing a Vision-Guided Sampler between the projector and the LLM, as well as a Text-Guided Sampler within the LLM. These works all perform token compression after the vision encoder.

Compared to previous works, our method differs in two key aspects: firstly, it selects the most pertinent visual tokens based on their relevance with the question and softly recycles less relevant ones using attention scores, minimizing information loss; secondly, it performs internal and hierarchical compression within the vision encoder, achieving lower computational costs at equivalent compression ratios.

\section{Methodology}
\subsection{Overview}
\begin{figure*}
\centering
    \includegraphics[width=0.98\textwidth]{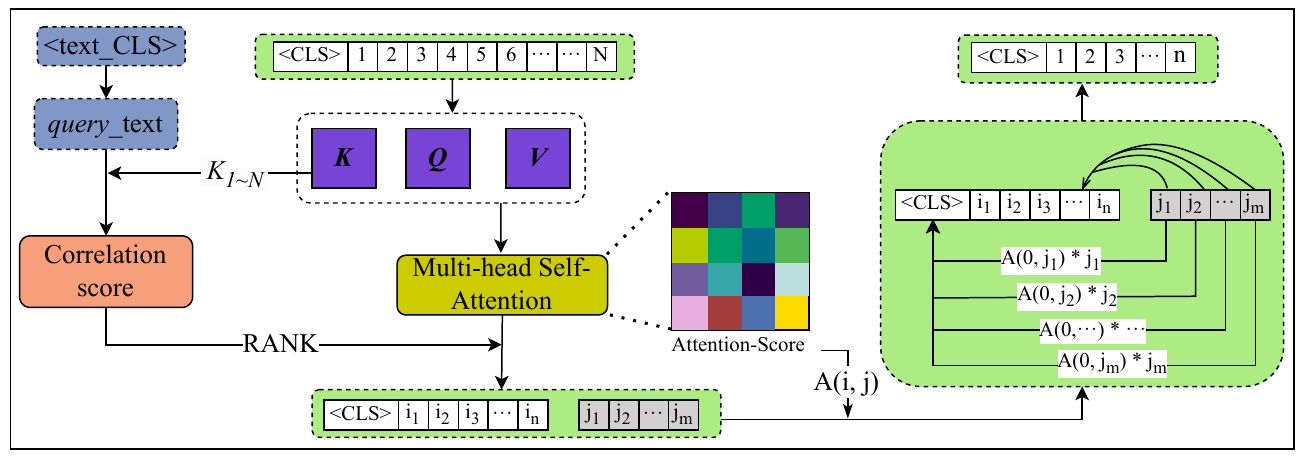}
\caption{Calculation details of the compression module.}
\label{fig2}
\end{figure*}

We propose \textit{\textbf{QG-VTC}}, which reduces visual tokens within the vision encoder through a hierarchical compression approach based on user's question. The overall architecture of our method is illustrated in Figure~\ref{fig1}, while the key compression module is described in detail in Figure~\ref{fig2}. The combination of original vision encoder and the compression modules constitutes the new vision encoder.

Our method essentially leverages textual tokens to perform fine-grained compression and selection of visual tokens across multiple intermediate layers of the original vision encoder. Both the original vision encoder and text encoder are derived from the CLIP model~\cite{radford2021learning}. This ensures that the textual information generated by the text encoder aligns more easily with the visual information produced by the vision encoder.

After hierarchical compression, the number of visual tokens is gradually reduced from $N$ to $M$, where $M$ is significantly less than $N$. The Projector maps the compressed visual tokens into the language space, and finally, the compressed visual information, along with the textual information from the user’s question, is fed into the LLM to generate an answer.

\subsection{Compression Module}
The key compression module is discribed in Figure~\ref{fig2}. For $N+1$ visual tokens (where 1 represents the $\textless CLS\textgreater{}$ token for the image), their $\textbf{\textit{Q}}$, $\textbf{\textit{K}}$, and $\textbf{\textit{V}}$ matrices are obtained through parameters $\boldsymbol{W}^{Q}$, $\boldsymbol{W}^{K}$, and $\boldsymbol{W}^{V}$, respectively. Subsequently, through Multi-head Self-Attention, an Attention-Score (A) matrix and $N+1$ new visual tokens are derived, as shown in Equation~\ref{eq1}\textasciitilde Equation~\ref{eq3}:
\begin{align}
\label{eq1}
\textbf{\textit{Q}} = Z \cdot \boldsymbol{W}^{Q}, \textbf{\textit{K}} = Z \cdot \boldsymbol{W}^{K}, \textbf{\textit{V}} = Z \cdot \boldsymbol{W}^{V}
\end{align}
\begin{align}
\label{eq2}
\text{A} &= \text{Softmax}\left(\frac{\textbf{\textit{Q}} \cdot \textbf{\textit{K}}^{T}}{\sqrt{d}}\right)
\end{align}
\begin{align}
\label{eq3}
Z' &= \text{A} \cdot \textbf{\textit{V}}
\end{align}
Where $Z$ denotes the original visual tokens, and $Z'$ denotes the derived visual tokens. Both $Z$ and $Z'$ $\in \mathbb{R}^{(N+1) \times d}$. $N$ denotes the number of visual tokens(except the $\textless CLS\textgreater{}$ token) and $d$ represents the dimension of each visual token.

The $\textless text\_CLS\textgreater{}$ token encapsulates the user's question information. Having been aligned to the image space through the MLP Layer, it can be directly projected into a query vector using the $\boldsymbol{W}^{Q}$. According to Equation~\ref{eq4}, this query vector is then used to compute the correlation score(C) with the $\textbf{\textit{K}}$ matrice of $N$ visual tokens, which reflects the relevance of each visual token to the user's question.
\begin{align}
\label{eq4}
\text{C} = \text{Softmax}\left(\frac{\mathbf{query} \cdot \mathbf{K}_{1:N}^\top}{\sqrt{d}}\right)
\end{align}

Based on the correlation score, the $N+1$ visual tokens are divided into two parts: the first part consists of the $\textless CLS\textgreater{}$ token and the top $n$ visual tokens with the highest correlation scores (the ones that need to be retained), while the second part comprises the tokens with the $m$ lowest correlation scores (the ones that need to be compressed).

To minimize the loss of global information, the compression module employs an Attention-Score(A) weighted averaging approach to incorporate the $m$ least significant visual tokens back into the $n+1$ retained tokens. Here, $A(i,j)$ represents the attention that $token_{i}$ pays to $token_{j}$. The information that a retained $token_{i}$ needs to reclaim from a discarded $token_{j}$ can be quantitatively expressed as $A(i,j) \times token_{j}$. The recycling process can be expressed through Equation~\ref{eq5}:
\begin{align}
\label{eq5}
token_{i} = token_{i} + \sum_{j=j_{1}}^{j_{m}} A(i,j) \times token_{j}
\end{align}
where $i \in (\textless CLS\textgreater{}, i_{1} \ldots i_{n})$. For brevity, the process can be described as $N$ visual tokens being compressed to $n$ by the compression module, disregarding the $\textless CLS\textgreater{}$ token.

\subsection{Vision Encoder}
CLIP-ViT/14 features a 24-layer Transformer architecture, and above compression module can be implemented based on any of these 24 layers. Several studies~\cite{ganz2024question} have indicated that the shallow layers of ViT are more focused on capturing low-level and local visual features, which exhibit weak correlations with semantics, whereas the deeper layers are more inclined to extract higher-level and more abstract visual features. Therefore, we have positioned the compression module within the deeper layers of ViT.

\begin{table}[htbp]
\centering
\resizebox{0.48\textwidth}{!}{
\begin{tabular}{cccccc}
\hline
Layer                & Question & Num. & Layer & Question & Num.  \\
\hline
0\textasciitilde{}11 & N     & 576  & ---   & ---   & ---   \\
$\textbf{12}$                   & Y     & 492  & 13    & N     & 492   \\
$\textbf{14}$                   & Y     & 408  & 15    & N     & 408   \\
$\textbf{16}$                   & Y     & 324  & 17    & N     & 324   \\
$\textbf{18}$                   & Y     & 240  & 19    & N     & 240   \\
$\textbf{20}$                   & Y     & 156  & 21    & N     & 156   \\
$\textbf{22}$                   & Y     & 72   & 23    & N     & 72    \\
\hline
\end{tabular}
}
\caption{Hierarchical Compression in Vision Encoder. "Layer" indicates the ViT layer index, with bold marking compressed layers. "Question" shows if the question ("Y" for yes, "N" for no) is used to guide compression. "Num." denotes the number of visual tokens output by each layer, excluding the <CLS> token.}
\label{tab1}
\end{table}

As discussed in FastV~\cite{chen2025image}, one transformer layer total Flops can be estimated by $4nd^{2} + 2n^{2}d + 2ndm$, where $n$ is the token number, $d$ is the hidden state size, and $m$ is the intermediate size of FFN. For the whole vision encoder, the theoretical Flops ratio $R$ is computed as Equation~\ref{eq6}:
\begin{align}
\label{eq6}
R=\frac{\sum_{i=0}^{23}\left ( 4\hat{n_{i}}d^{2}+2\hat{n_{i}}^{2}d+2\hat{n_{i}}dm \right ) }{24\left ( 4nd^{2}+2n^{2}d+2ndm \right )}
\end{align}
where $\hat{n_{i}}$ is the visual token number in the $i_{th}$ layer.

In previous compression schemes, visual information typically undergoes a one-time, abrupt compression process, which often results in the loss of critical information, thereby impacting the performance of subsequent vision-related tasks. To ensure the smoothness of the compression process, the compression module should not reduce the number of visual tokens too abruptly in a single step.

Therefore, within the deeper layers of ViT, we adopt a hierarchical compression approach, which can be understood as a gradual compression method. Each layer of compression further refines the information based on the previous layer. This progressive approach helps mitigate the sharp loss of information that occurs in single-step compression.

More importantly, by incorporating hierarchical compression, we can significantly reduce the computational load of the visual encoder itself, and consequently enhancing overall computational efficiency.

When the image resolution is $336 \times 336$, the detailed information regarding the hierarchy and the number of compressions is presented in Table~\ref{tab1}. Subsequent ablation experiments have also demonstrated the effectiveness of this hierarchical compression approach.

\section{Experiments}
\subsection{Datasets and Metrics}\label{sec:4.1}
To assess the capabilities of our method, we conduct comprehensive experiments on multiple benchmarks, including VQA\textsuperscript{v2}~\cite{goyal2017making}, GQA~\cite{hudson2019gqa}, VisWiz~\cite{gurari2018vizwiz}, SQA\textsuperscript{I}~\cite{lu2022learn}, VQA\textsuperscript{T}~\cite{singh2019towards}, and POPE~\cite{li2023evaluating}. For more detailed information on these datasets and metrics, please refer to Appendix~\ref{appendix: A}.

All results are evaluated according to the official metrics of the datasets. Additionally, the results for the VQA\textsuperscript{v2} and VisWiz are obtained by uploading our predicting outputs to the official challenge website.

\subsection{Implementation Details}\label{sec:Implementation Details}
We adhere to the two-stage training strategy, where we first pretrain the Projector and the MLP Layer depicted in Figure~\ref{fig1} using the LCS-558K subset~\cite{liu2024improved}. Subsequently, we fine-tune both the Projector and the LLM(LoRA) using the LLaVA-Instruct-665K dataset~\cite{liu2024improved}. Notably, we adapted the LLaVA-Instruct-665K dataset to ensure that each image corresponds to only one question, implying that the model needs to re-select visual features in response to each single question provided. For more detailed information on these datasets, please refer to Appendix~\ref{appendix: A}, and detailed information regarding the training process is provided in Appendix~\ref{appendix: B}.
\begin{table*}[htb]
\centering
\arrayrulecolor{black}
\begin{tabular}{>{\centering\hspace{0pt}}m{0.16\linewidth}>{\centering\hspace{0pt}}m{0.08\linewidth}>{\centering\hspace{0pt}}m{0.08\linewidth}>{\centering\hspace{0pt}}m{0.08\linewidth}>{\centering\hspace{0pt}}m{0.08\linewidth}>{\centering\hspace{0pt}}m{0.08\linewidth}>{\centering\hspace{0pt}}m{0.08\linewidth}>{\centering\arraybackslash\hspace{0pt}}m{0.08\linewidth}}
\arrayrulecolor{black}\hline
Method                         & N & VQA\textsuperscript{v2} & GQA           & VisWiz         & SQA\textsuperscript{I} & VQA\textsuperscript{T} & POPE            \\
\cline{1-1}\arrayrulecolor{black}\cline{2-8}
Baseline   & 576   & 78.5  & 62  & 50  & 66.8 & 58.2 & 85.9  \\
\hline
CrossGET                       & 288   & 77.3  & ---  & ---  & 66.7  & 54.9  & 83.90  \\
Prumerge+     & 144   & 76.8    & ---  & ---   & 68.3   & 57.1  & 84.0   \\
FastV        & 144   & 74.07   & 56.58 & 51.29  & 69.11   & 57.38  & 73.74  \\
FitPrune     & 144   & 76.14   & 59.38  & 51.30  & 69.01  & 56.49  & 80.75   \\
SparseVLM    & 144   & 72.76   & 55.11  & \textbf{51.46} & 69.36   & 55.99  & 77.57    \\
DeCo        & 144   & 74.0     & 54.1   & 49.7       & ---    & 56.2   & 84.6   \\
\textbf{Ours} & 144   & \textbf{77.67}  & \textbf{60.5} & 48.72  & \textbf{69.39}  & \textbf{57.90}  & \textbf{85.21}  \\
\hline
VisionZip & 128    & 76.6  & 58.9  & ---   & 68.3  & 57.0  & 83.7     \\
Trim      & 121    & 76.4  & \textbf{61.4} & 48.1  & 69.1  & 53.7  & \textbf{85.3}    \\
\textbf{Ours} & 120    & \textbf{76.96}  & 56.69      & \textbf{48.84} & \textbf{69.39} & \textbf{57.42} & 84.58   \\
\hline
              & 96    & 76.08 & 59.4 & 48.14  & 69.61 & 57.52  & 84.48  \\
\textbf{Ours} & 72    & 75.31 & 58.49 & 51.05  & 69.79 & 57.44  & 82.32  \\
              & 36    & 71.84 & 56.61 & 50.86  & 69.98 & 54.98  & 80.55  \\
\hline
\end{tabular}
\arrayrulecolor{black}
\caption{Comparison with existing SOTA vision token compression or prune methods on different benchmarks. "Baseline" denotes the performance of LLaVA-1.5-7B without visual token compression, "N" denotes the number of visual tokens after compression, and \textbf{Bold} means the best performance.}
\label{tab2}
\end{table*}

\subsection{Main Results}\label{sec:4.3}
Our experiments builds upon LLaVA-1.5-7B~\cite{liu2024improved} and focuses on the issue of visual token compression. For fair and consistent comparison, we compare \textbf{\textit{QG-VTC}} with state-of-the-art(SOTA) vision token compression or prune methods, including CrossGET~\cite{shi2023crossget}, Prumerge+~\cite{shang2024llava}, FastV~\cite{chen2025image}, SparseVLM~\cite{zhang2024sparsevlm}, DeCo~\cite{yao2024deco}, VisionZip~\cite{yang2024visionzip}, Trim~\cite{song2024less}, which are also based on LLaVA-1.5-7B.

As illustrated in Table~\ref{tab2}, when the final vision tokens are compressed to 144, QG-VTC attains SOTA performance across 5 benchmarks. When the final number is compressed to 120, although slightly fewer than those of VisionZip and Trim, QG-VTC still achieves SOTA results on 4 benchmarks.

To explore the potential of QG-VTC, the number of vision tokens is further compressed to 96, 72, and 36. Overall, as the number decreases, the performance exhibits a downward trend. This can be interpreted as the more compact the visual information is compressed, the greater the difficulty for the model in decompressing it when understanding visual features. When the number is 72, its performance can be maintained at above 94.3\% of the original performance. Further, when the number of vision tokens is reduced to 36, its performance can only be maintained at above 91.3\%.
\begin{figure}
\centering
    \includegraphics[width=0.49\textwidth]{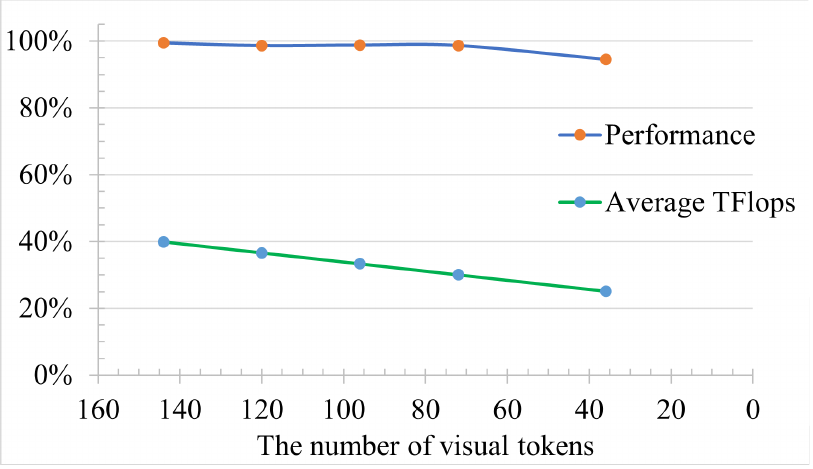}
\caption{The relationship between computational load and performance (evaluation on VQA\textsuperscript{T}). 100\% means the baseline's (576 visual tokens) performance and computational load.}
\label{fig3}
\end{figure}

Taking the VQA\textsuperscript{T} dataset as an example, we investigates the relationship between performance and computational load during the compression process, using the original model as the baseline. 10 samples are randomly selected to compute the average computational load during inference. Please refer to Appendix~\ref{appendix: C} for more detailed calculation information.

As illustrated in Figure~\ref{fig3}, the average computational load(quantified by TFlops) decreases almost linearly with the reduction in the number. For instance, when the number is 72, QG-VTC maintains 98.69\% of the baseline performance, while the average computational load is only 30\% of the baseline. As the number further decreases, performance declines more rapidly. Considering the trade off between performance and computational efficiency, compressing the number to 72 represents a better choice. Please refer to Appendix~\ref{appendix: B} for more detailed Visualization information.

\begin{table*}[htb]
\centering
\arrayrulecolor{black}
\resizebox{0.95\textwidth}{!}{
\begin{tabular}{cccccccc}
\hline
Method & Layer      &R    &Question       &Recycle               & GQA            & VQA\textsuperscript{T} & POPE            \\
\hline
A1     & 1-3-5-7-9-11      &36.87\%    &\Checkmark &\Checkmark      & 54.83          & 47.89                  & 80.22           \\
A2     & 12-13-14-15-16-17 &68.46\%    &\Checkmark &\Checkmark      & 56.81          & 55.01                  & 82.07           \\
A3     & 17-18-19-20-21-22 &86.88\%    &\Checkmark &\Checkmark      & 57.32          & 54.48                  & 81.97           \\
\hline
A4     &12                 &59.47\%    &\Checkmark &\Checkmark      & 53.48          &50.39           & 77.76 \\
A5     &16                 &74.21\%    &\Checkmark &\Checkmark      & 53.31          &53.61           & 79.18 \\
A6     &20                 &88.95\%    &\Checkmark &\Checkmark      & 57.16          &55.39           & 81.22 \\
\hline
A7     & 12-14-16-18-20-22 &77.36\%    &           &                & 56.31           & 55.18                 & 80.43            \\
A8     & 12-14-16-18-20-22 &77.36\%    &           &\Checkmark      & 56.68           & 55.85                  & 81.27           \\
A9     & 12-14-16-18-20-22 &77.40\%    &\Checkmark &        &57.25           & 56.06                  & 81.41           \\
\hline
QG-VTC   & 12-14-16-18-20-22 &77.40\%  &\Checkmark &\Checkmark      & \textbf{58.49} & \textbf{57.44}         & \textbf{82.32}  \\
\hline
\end{tabular}
\arrayrulecolor{black}
}
\caption{Comparison of different compression architectures. "Layer" indicates the location of the compression module. "Question" indicates whether the compression process is guided by the user's question, "Recycle" indicates whether the attention score weighted averaging approach described in Equation~\ref{eq3} is utilized in the compression process, and "R" indicates the computational load ratio of the vision encoder.}
\label{tab3}
\end{table*}
\subsection{Visualization}\label{sec:4.4}

To gain a clearer understanding of which vision tokens are retained and which are compressed based on user's question, we show visualization results on some samples. In these experiments, each image's visual tokens were compressed to 72 tokens (1/8 of the original number of visual tokens) according to the hierarchical compression structure outlined in Table~\ref{tab1}.
\begin{figure}[htb]
    \centering
    \begin{minipage}{\columnwidth}
        \centering
        \begin{subfigure}[b]{0.493\columnwidth}
            \centering
            \includegraphics[width=\textwidth]{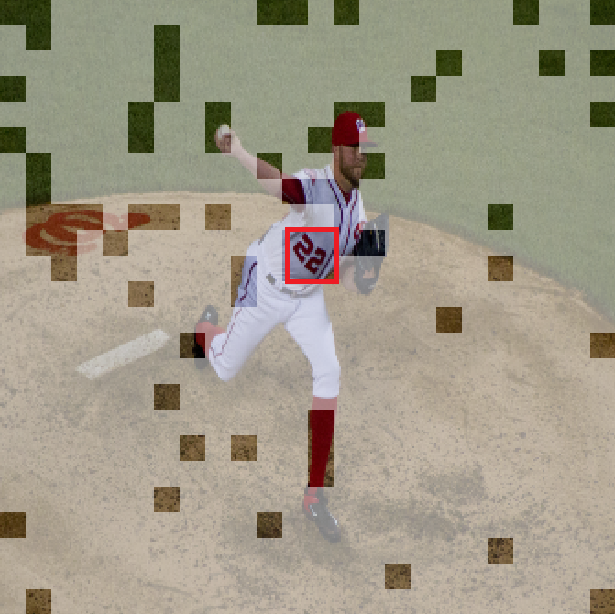}
        \end{subfigure}
        \hfill
        \begin{subfigure}[b]{0.493\columnwidth}
            \centering
            \includegraphics[width=\textwidth]{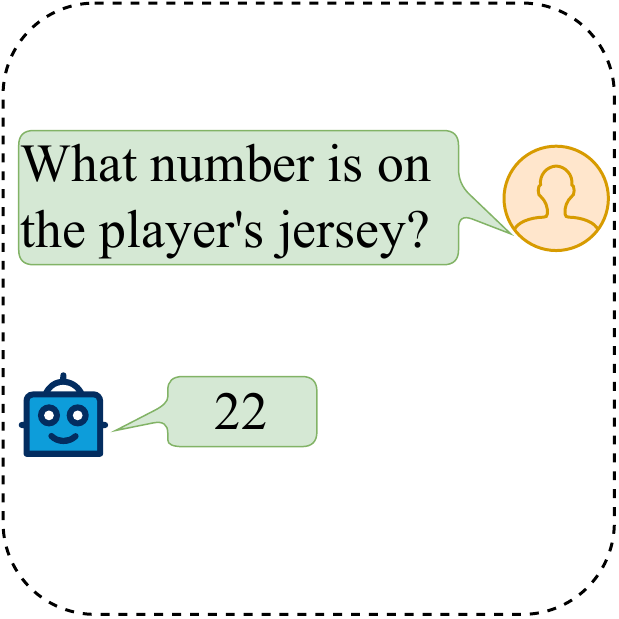}
        \end{subfigure}
        \begin{subfigure}[b]{0.493\columnwidth}
            \centering
            \includegraphics[width=\textwidth]{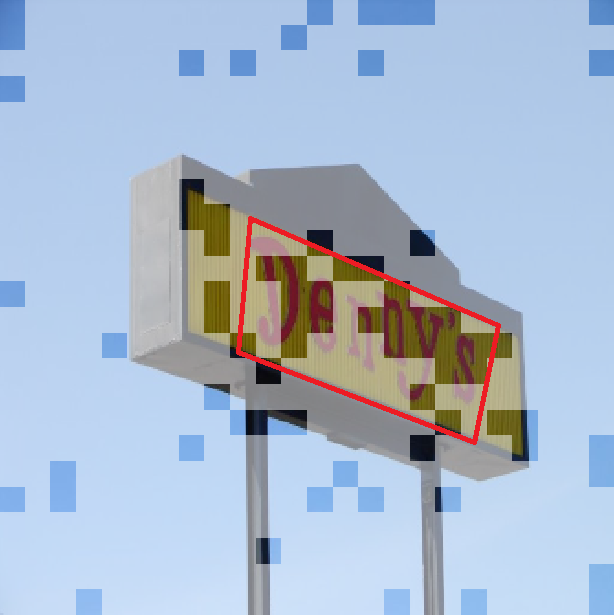}
        \end{subfigure}
        \hfill
        \begin{subfigure}[b]{0.493\columnwidth}
            \centering
            \includegraphics[width=\textwidth]{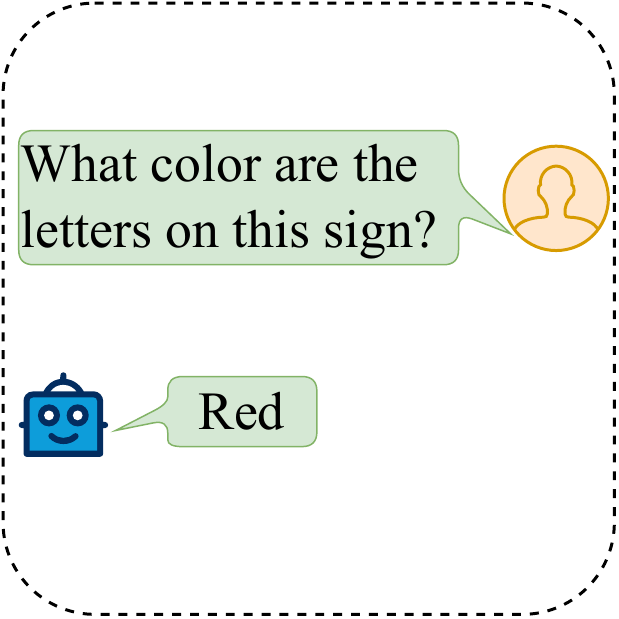}
        \end{subfigure}
        \caption{The visualization results of \textit{\textbf{QG-VTC}}. {\color{red}The red box} represents the area corresponding to the answer. The unmasked areas indicate the retained visual tokens.}
        \label{fig4}
    \end{minipage}
\end{figure}
As shown in the Figure~\ref{fig4}, the masked tokens represent those that are compressed, indicating their lower relevance to the user's question. In the sample illustrated in the first row, visual tokens inside the red box which are closely related to the question are fully preserved. Our method effortlessly provides the correct answer. In the second row sample, though not all tokens inside the red box are preserved, our model can still accurately answer the user's question based on the preserved visual tokens.

Although the compression process may filter out some crucial visual information, the attention-score weighted averaging approach integrates the discarded information into the preserved visual tokens. Subsequently, relying on the powerful decoding capability of the LLM, this information can potentially be re-decoded, aiding the LLM in responding to the user's question.

The ablation studies conducted in Section~\ref{sec: Component-wise Analysis} also validate the significance of the attention-score weighted averaging approach. Please refer to Appendix~\ref{appendix: D} for more detailed visualization information.

\subsection{Ablation Studies}
By inserting the compression module from Figure~\ref{fig2} into different layers of the vision encoder, we explore the impact of various compression architectures. All architectures compress the visual tokens to 72, with experimental results presented in Table~\ref{tab3}. $R$ indicates the computational load ratio of the vision encoder to that of the original vision encoder (without compression), which is theoretically calculated via Equation~\ref{eq6}.

\subsubsection{Compression Architectures}

Specifically, A1 represents placing the compression module in the early layers of the vision encoder. Its suboptimal performance demonstrates that low-level visual features exhibit weak correlations with semantics. At this stage, semantic features are unable to effectively pick out the relevant visual features.

Both A2 and A3 perform compression in deeper layers, but unlike QG-VTC, they both adopt a continuous compression approach, which lacks an intermediate layer between two contiguous compression modules. An intermediate layer allows for a new interaction among the compressed visual information, thereby enhancing the smoothness of the entire compression process. These experiments indicate that such interaction is beneficial.

In experiments A4 to A6, we modify the hierarchical compression approach to a single-step method, where visual tokens are directly compressed from 576 to 72. The results show that later single-step compression improves model performance but increases computational cost. In contrast, multi-step compression can achieve better performance with a lower computational load than single-step methods.

From the perspective of balancing the computational load and performance of the vision encoder, the hierarchical compression architecture presented in Table~\ref{tab1} represents the optimal choice.

\subsubsection{Component-wise Analysis}\label{sec: Component-wise Analysis}
To investigate the role of user's question and visual information recycling mechanisms in the compression process, we design experiments A7, A8, and A9.

In A7, rather than utilizing user's question as guidance during the compression process, the $\textless CLS\textgreater{}$ token of the image itself is employed to calculate the correlation score as illustrated in Figure~\ref{fig2}. This correlation score is then used to rank the visual tokens and the lower-ranked are directly discarded. In this scenario, the compression process can be regarded as a pure pruning process.

A8 builds upon A7 by incorporating the attention score weighted averaging approach to incorporate the discarded visual tokens into the retained ones. A9 indicates that user's question is utilized as guidance during the compression process, but similar to A4, it remains a purely pruning procedure.

As demonstrated by these experiments, both "Question" and "Recycle" play crucial roles in compression process, thereby validating the effectiveness and rationality of QG-VTC.

\section{Conclusion}
In this work, we propose QG-VTC, an innovative approach for visual token compression. Guided by the user's question, this method not only identify and select visual information highly relevant to the user's question but also recycle the less relevant information through an attention score weighted averaging approach. We also propose a multi-step strategy that can conduct the above visual token compression operation progressively and hierarchically. Consequently, this achieves a reduction in the number of visual tokens within the vision encoder itself. Experimental results demonstrate that QG-VTC maintains at least 97.44\% or 94.3\% performance across multiple datasets when using only 1/4 or 1/8 of the original vision tokens. From the perspective of computational efficiency, while maintaining comparable performance to the original, QG-VTC can reduce the computational load of the visual encoder and the overall model to 77\% and 30\% of the original, respectively. We aspire to apply our methodology in the construction of a document-based intelligent question-answering system, which highly relies on high resolution visual information and will be the focus of our future research endeavors.
\section*{Limitations}
There remain several limitations in our methodology. Firstly, our approach compresses visual information using a predefined compression rate. However, in practical applications, we have observed that for certain tasks, less refined visual information may be sufficient or alternatively, additional visual information may be necessary to provide satisfactory answers. To address this shortcoming, our next step will be to investigate the use of dynamic compression rates for compressing visual information. Secondly, constrained by hardware and computational resources, we have not yet discussed the scenario involving high-resolution inputs,especially for document related VQA tasks, which may necessitate fine-tuning of all parameters within the vision encoder.

\section*{Ethical Impact}
We respect intellectual property rights and comply with relevant laws and regulations. The datasets in our paper are publicly available, and we have taken careful measures to ensure that our research does not contain any personal sensitive information. In addition. our work is only for research purposes, not for commercial purposes.

\section*{Acknowledgments}

This information will be submitted later.

\bibliography{custom}

\appendix
\section{Datasets Details}
\label{appendix: A}
The detailed information of the evaluation benchmarks mentioned in section~\ref{sec:4.1} and section~\ref{sec:Implementation Details} is provided below:

VQA\textsuperscript{v2}~\cite{goyal2017making} is a more balanced VQA dataset, with significantly reduced language biases. When answering the questions, models need to focus more on visual information. Its test set contains more than 453K (question, image) pairs and is divided into 4 splits: test-dev, test-standard, test-challenge and test-reserve. The accuracy is computed by using 10 ground-truth answers for each question. In our work, the test-standard accuracy is used to evaluate the model’s visual perception capabilities.

GQA~\cite{hudson2019gqa} centers around real-world reasoning, scene understanding and compositional question answering. All images, questions and answers are accompanied by matching semantic representations. The answer distribution for each question group is more uniform than VQAv2, which makes the educated guesses strategy far less rewarding and demands instead more refined comprehension of both the visual and linguistic contents. In our work, the test-dev accuracy is used to evaluate the model’s behavior and performance.

VizWiz~\cite{gurari2018vizwiz} is the first publicly-available vision dataset which originates from a natural use case where blind people took images and then asked questions about them. Images are often poor quality (e.g., blur, poor lighting), and questions cannot be assured to have answers. It is difficult for modern models. In our work, the test-dev accuracy is computed to evaluate the model’s performance.

SQA\textsuperscript{I}~\cite{lu2022learn} is a multimodal multiple-choice science question dataset. It covers diverse topics across natural science, social science, and language science subjects. Most questions are annotated with grounded lectures and detailed explanations. To effectively answer the questions, a model often needs to be able to understand the multimodal content in the input and extract external knowledge, similar to how humans do. In our work, the test accuracy is computed to evaluate the model’s performance.

VQA\textsuperscript{T}~\cite{singh2019towards} contains questions asked by humans on images from Openimages~\cite{2016OpenImages}, as well as the OCR information extracted from each image. To answer these questions, models need to not only parse the image and the question but also read the text in the image, identify which text might be relevant to the question, and further recognize whether any subset of the detected text can directly serve as the answer. Each question-image pair is associated with 10 human-provided answers, which serve as the ground truth for evaluation. The model's capabilities are assessed using the accuracy on the validation set.

POPE~\cite{li2023evaluating} formulates the evaluation of object hallucination as a binary classification task that prompts MLLMs to output “Yes” or “No”. By devising random, popular and adversarial sampling strategies on the validation set of MSCOCO~\cite{lin2014microsoft}, POPE can validate whether models are prone to hallucinate specific objects. The average accuracy of 3 sampling strategies is computed to evaluate the model’s performance.

LCS-558K~\cite{liu2024improved} is a subset of 558K image-text pairs from LAION-CC-SBU with BLIP captions~\cite{pmlr-v162-li22n}, which has been extensively used in the LLaVA series. During the pretraining process, the text component of each image-text pair can be regarded as an answer, while the question is derived from a predefined list of instructions aimed at providing brief descriptions of images, e.g., `Provide a brief description of the given image.', `Give a short and clear explanation of the subsequent image.', `Create a compact narrative representing the image presented.' and others.

LLaVA-Instruct-665K dataset~\cite{liu2024improved} is a mixed visual instruction tuning dataset, which contains a variety of datasets. All data splits are concatenated together and sampled with the same probability.

\section{Implementation Details}
The detailed information of the training process mentioned in section~\ref{sec:Implementation Details} is presented in Table~\ref{tab4}. During the pretraining stage, the learning rate for the Projector and the MLP Layer is set to 5e-4. During the fine-tuning stage, the learning rate for the Projector and the MLP Layer is set to 1e-5 and the learning rate for the LLM(utilizing LoRA) is set to 1e-4. Both the pretraining and fine-tuning process are implemented on 2 NVIDIA A6000 GPUs.
\label{appendix: B}
\begin{table}[htbp]
\centering
\arrayrulecolor{black}
\begin{tabular}{cclc}
\hline
Settings                  & \multicolumn{2}{c}{Stage1} & Stage2  \\
\hline
Batch size(single device) & \multicolumn{2}{c}{64}     & 16      \\
Learning rate1            & \multicolumn{2}{c}{5e-4}   & 1e-5    \\
Learning rate2(LoRA)            & \multicolumn{2}{c}{---}    & 1e-4    \\
LoRA\_R                   & \multicolumn{2}{c}{---}    & 256     \\
LoRA\_$\alpha$                   & \multicolumn{2}{c}{---}    & 128     \\
Learning schedule         & \multicolumn{3}{c}{Cosine decay}     \\
Warmup ratio              & \multicolumn{3}{c}{0.03}             \\
Weight decay              & \multicolumn{3}{c}{0}                \\
Epoch                     & \multicolumn{3}{c}{1}                \\
Optimizer                 & \multicolumn{3}{c}{AdamW}            \\
DeepSpeed stage           & \multicolumn{3}{c}{2}                \\
MLP Layer                 & \multicolumn{3}{c}{Open}             \\
LLM                       & Freeze & \multicolumn{2}{c}{LoRA}    \\
Vision\_encoder           & \multicolumn{3}{c}{Freeze}           \\
Text\_encoder             & \multicolumn{3}{c}{Freeze}           \\
Projector                 & \multicolumn{3}{c}{Open}             \\
Max token                 & \multicolumn{3}{c}{2048}             \\
\arrayrulecolor{black}\hline
\end{tabular}
\caption{Training settings of QG-VTC.}
\label{tab4}
\end{table}

\section{Computational Load Information}
The detailed information of the computational load mentioned in section~\ref{sec:4.3} is presented in Table~\ref{tab5}. We randomly select 10 samples from the VQA\textsuperscript{T}~\cite{singh2019towards} dataset and compute the average computational load, quantified in TFlops, during the inference phase. $N=576$ means the baseline. This table offers a more direct illustration of why compressing the number of visual tokens to 72 represents a favorable trade off, effectively balancing performance and computational efficiency.
\label{appendix: C}
\begin{table*}
\centering
\resizebox{0.99\textwidth}{!}{
\begin{tabular}{cccccccccccccc}
\hline
\multicolumn{1}{c|}{N} & \multicolumn{10}{c|}{TFLOPs}                                         & Ave  & ratio &Acc\% \\
\hline
\multicolumn{1}{c|}{576} & 9.11 & 9.52 & 11.01 & 11.04 & 8.98 & 8.92 & 8.95 & 11.07 & 9.10 & \multicolumn{1}{c|}{8.99} & 9.67  & 100\% &100\%                      \\
\hline
\multicolumn{1}{c|}{144} & 3.30 & 3.71 & 5.20  & 5.23  & 3.17 & 3.12 & 3.14 & 5.26  & 3.29 & \multicolumn{1}{c|}{3.18} & 3.86  & 39.92\% &99.48\%                    \\
\multicolumn{1}{c|}{120} & 2.98 & 3.39 & 4.88  & 4.91  & 2.85 & 2.80 & 2.82 & 4.94  & 2.97 & \multicolumn{1}{c|}{2.86} & 3.54  & 36.61\% &98.66\%                   \\
\multicolumn{1}{c|}{96}  & 2.66 & 3.07 & 4.57  & 4.59  & 2.53 & 2.48 & 2.50 & 4.62  & 2.65 & \multicolumn{1}{c|}{2.54} & 3.22  & 33.31\% &98.83\%                  \\
\multicolumn{1}{c|}{72}  & 2.34 & 2.75 & 4.25  & 4.27  & 2.21 & 2.16 & 2.18 & 4.30  & 2.33 & \multicolumn{1}{c|}{2.22} & 2.90  & 30.00\% &98.69\%                 \\
\multicolumn{1}{c|}{36}  & 1.86 & 2.27 & 3.77  & 3.79  & 1.73 & 1.68 & 1.71 & 3.82  & 1.85 & \multicolumn{1}{c|}{1.75} & 2.42  & 25.06\% &94.47\%                   \\
\hline
\end{tabular}
}
\caption{Computational Load on a single NVIDIA A6000 GPU. "N" denotes the final number of visual tokens outputted by the vision encoder. "Ave" means the average computational load of the ten samples. "Acc\%" represents the ratio of the performance achieved by various N to that of the original model, across the entire VQA\textsuperscript{T} dataset.}
\label{tab5}
\end{table*}

\section{Visualization Information}
\begin{figure*}[h]
    \centering
    \begin{subfigure}[b]{0.3\textwidth}
        \centering
        \includegraphics[width=\textwidth]{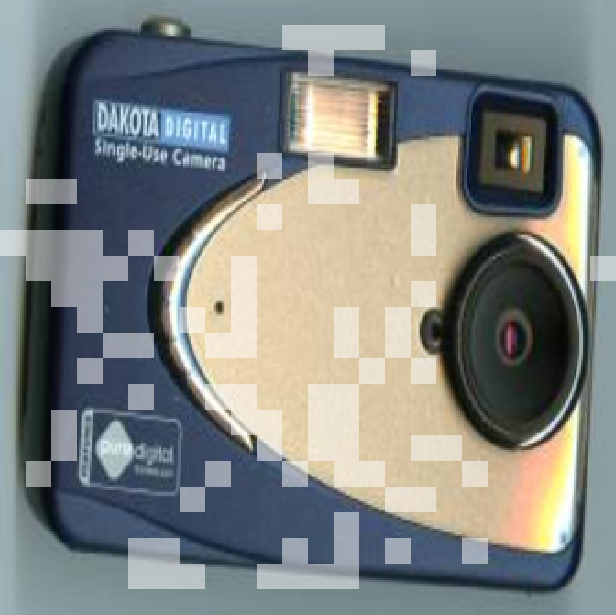}
        \caption{492}
    \end{subfigure}
    \hfill
    \begin{subfigure}[b]{0.3\textwidth}
        \centering
        \includegraphics[width=\textwidth]{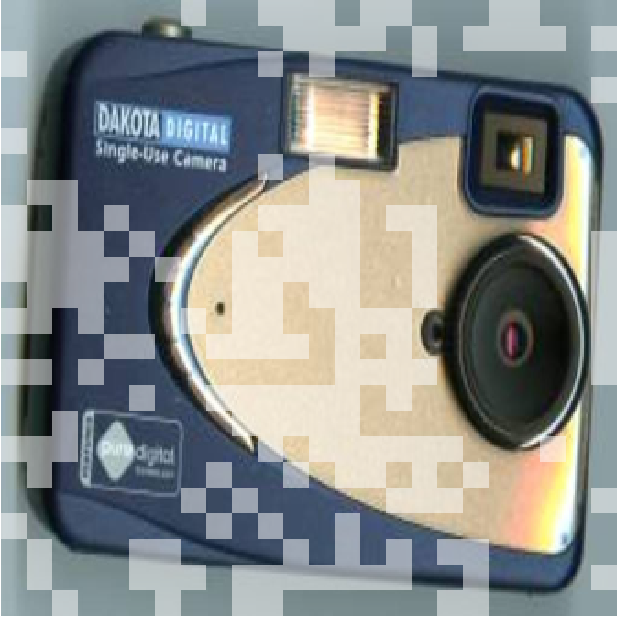}
        \caption{408}
    \end{subfigure}
    \hfill
    \begin{subfigure}[b]{0.3\textwidth}
        \centering
        \includegraphics[width=\textwidth]{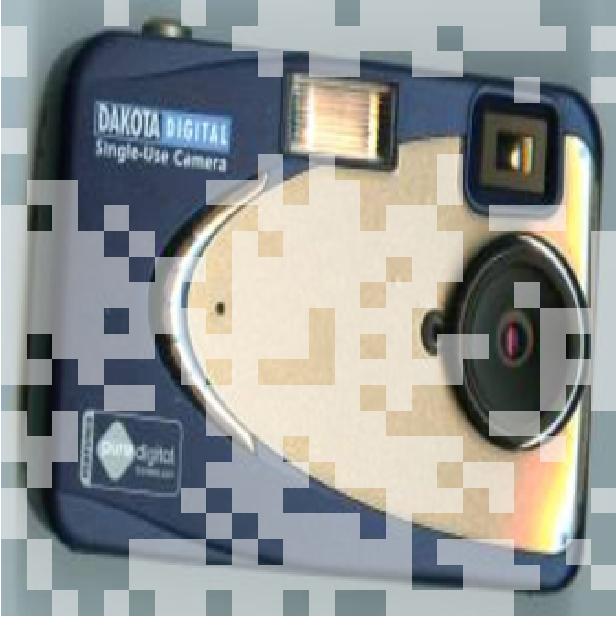}
        \caption{324}
    \end{subfigure}

    \begin{subfigure}[b]{0.3\textwidth}
        \centering
        \includegraphics[width=\textwidth]{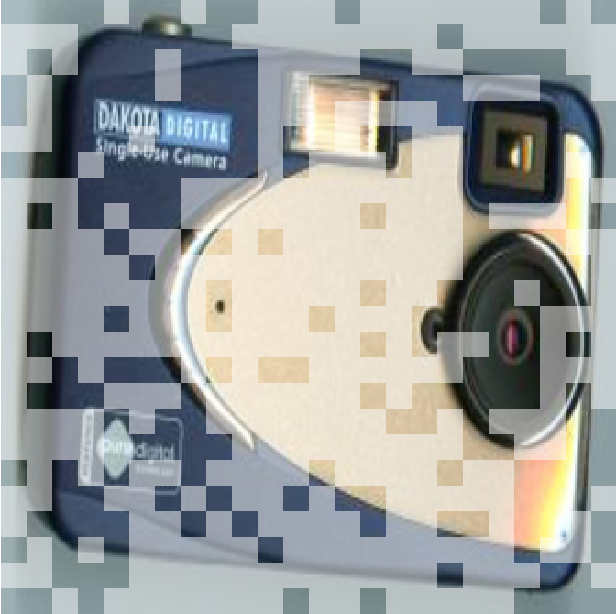}
        \caption{240}
    \end{subfigure}
    \hfill
    \begin{subfigure}[b]{0.3\textwidth}
        \centering
        \includegraphics[width=\textwidth]{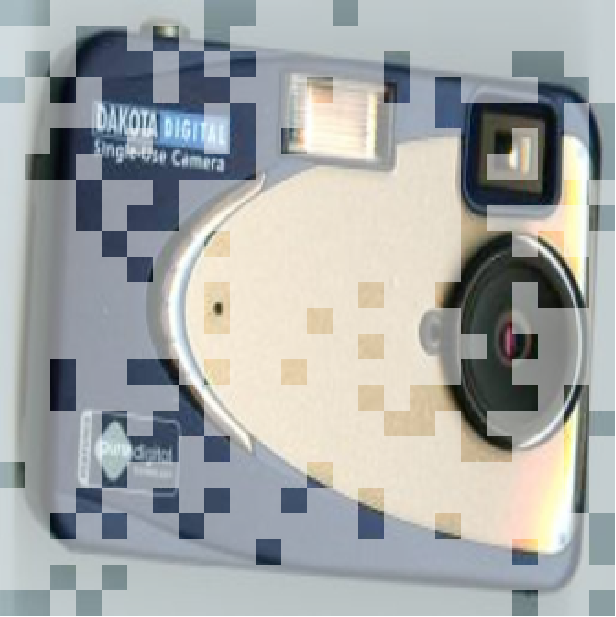}
        \caption{156}
    \end{subfigure}
    \hfill
    \begin{subfigure}[b]{0.3\textwidth}
        \centering
        \includegraphics[width=\textwidth]{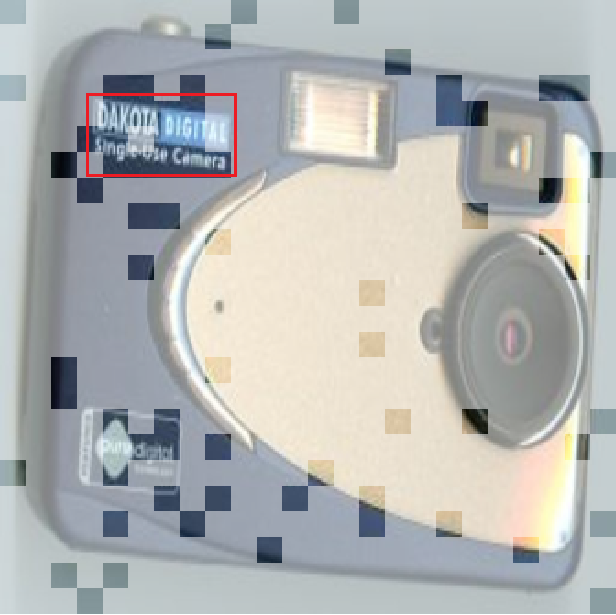}
        \caption{72}
    \end{subfigure}

    \caption{\textbf{Question}: What is the brand of this camera? {\color{blue}Ground\_truth}: dakota digital. {\color{green}Output}: dakota digital.(\Checkmark) 492, 408, 324, 240, 156, 72 represent the number of retained visual tokens, respectively. {\color{red}The red box} represents the visual tokens corresponding to the answer.}
    \label{fig5}
\end{figure*}

More visual token compression visualization results are presented in Figure~\ref{fig5}\textasciitilde Figure~\ref{fig9}. From these results we can see that as hierarchical compression progresses, the model gradually focuses on the most relevant visual regions. It is undeniable that there still exist some visual information irrelevant to the answer, which can be regarded as high-frequency noise. We analyze that this is due to the fact that, after the model identifies the most relevant visual regions, the number of vision tokens falls below the prescribed compression rate, resulting in some extraneous visual information being included as fillers. This also underscores the necessity of exploring dynamic compression rates in our future endeavors.
\label{appendix: D}

\begin{figure*}[h]
    \centering

    \begin{subfigure}[b]{0.3\textwidth}
        \centering
        \includegraphics[width=\textwidth]{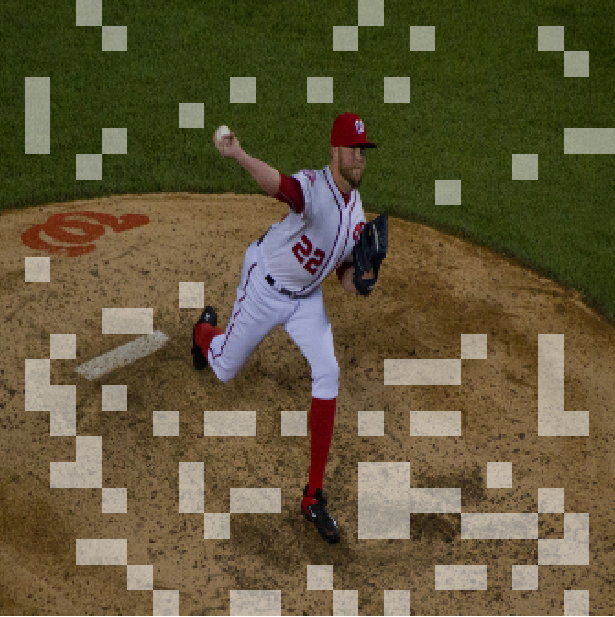}
        \caption{492}
    \end{subfigure}
    \hfill
    \begin{subfigure}[b]{0.3\textwidth}
        \centering
        \includegraphics[width=\textwidth]{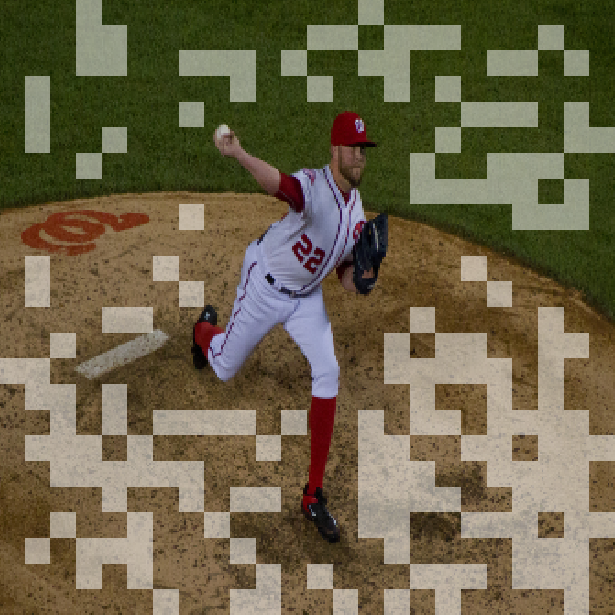}
        \caption{408}
    \end{subfigure}
    \hfill
    \begin{subfigure}[b]{0.3\textwidth}
        \centering
        \includegraphics[width=\textwidth]{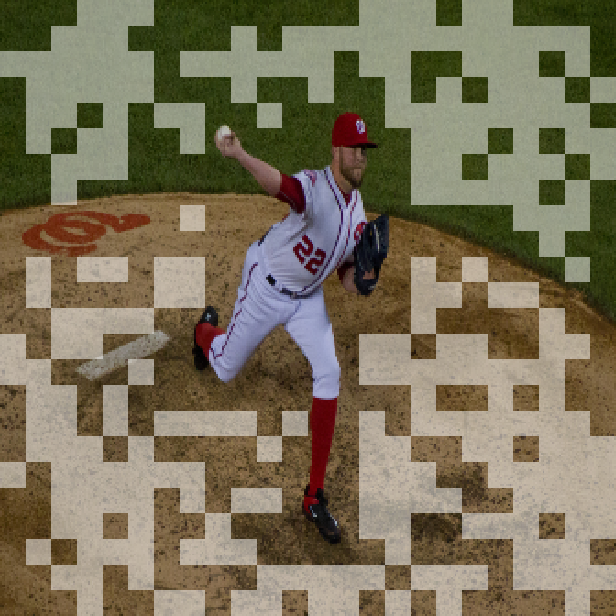}
        \caption{324}
    \end{subfigure}

    \begin{subfigure}[b]{0.3\textwidth}
        \centering
        \includegraphics[width=\textwidth]{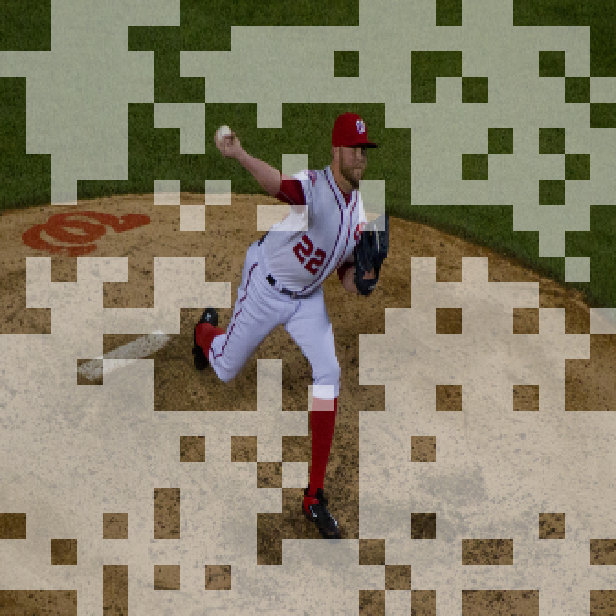}
        \caption{240}
    \end{subfigure}
    \hfill
    \begin{subfigure}[b]{0.3\textwidth}
        \centering
        \includegraphics[width=\textwidth]{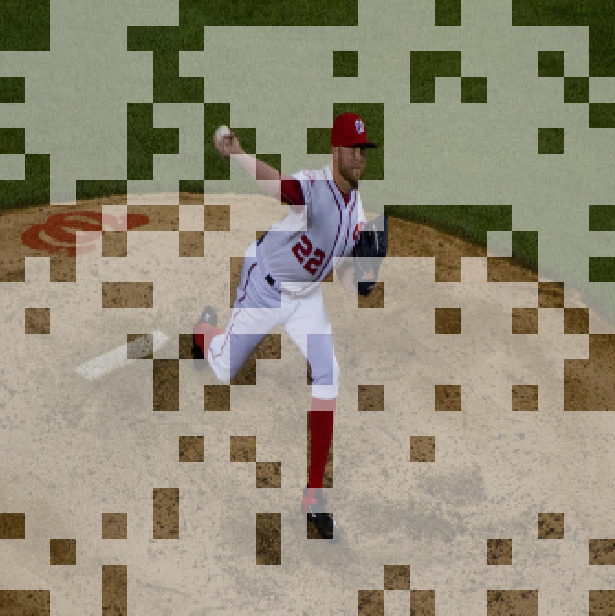}
        \caption{156}
    \end{subfigure}
    \hfill
    \begin{subfigure}[b]{0.3\textwidth}
        \centering
        \includegraphics[width=\textwidth]{./visual/94ad4aad01e27a32/72.png}
        \caption{72}
    \end{subfigure}

    \caption{\textbf{Question}: What number is on the player's jersey? {\color{blue}Ground\_truth}: 22. {\color{green}Output}: 22.(\Checkmark)}
    \label{fig6}
\end{figure*}

\begin{figure*}[h]
    \centering

    \begin{subfigure}[b]{0.3\textwidth}
        \centering
        \includegraphics[width=\textwidth]{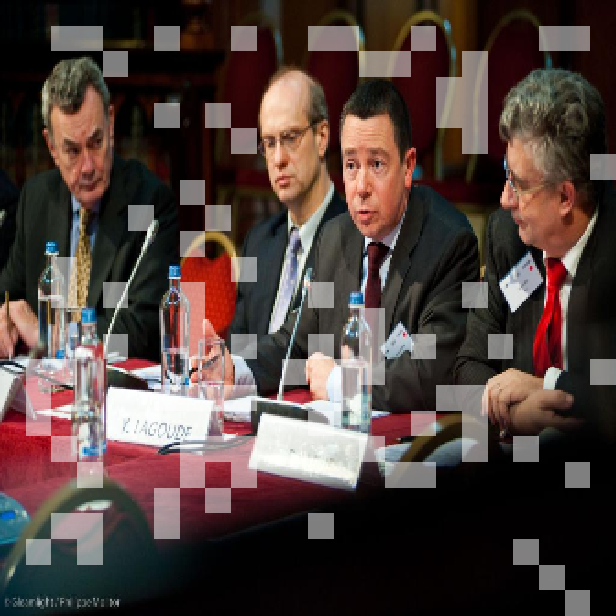}
        \caption{492}
    \end{subfigure}
    \hfill
    \begin{subfigure}[b]{0.3\textwidth}
        \centering
        \includegraphics[width=\textwidth]{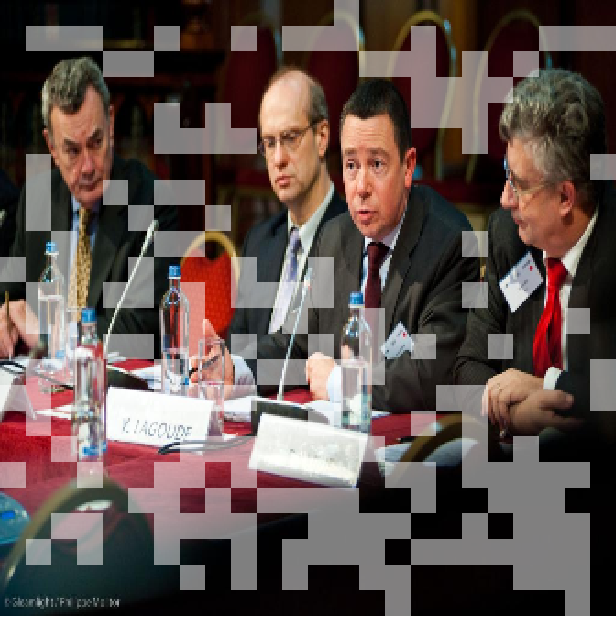}
        \caption{408}
    \end{subfigure}
    \hfill
    \begin{subfigure}[b]{0.3\textwidth}
        \centering
        \includegraphics[width=\textwidth]{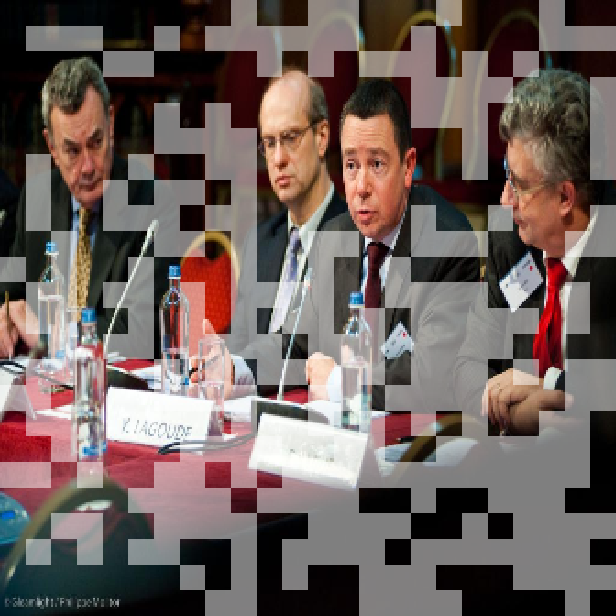}
        \caption{324}
    \end{subfigure}
    \begin{subfigure}[b]{0.3\textwidth}
        \centering
        \includegraphics[width=\textwidth]{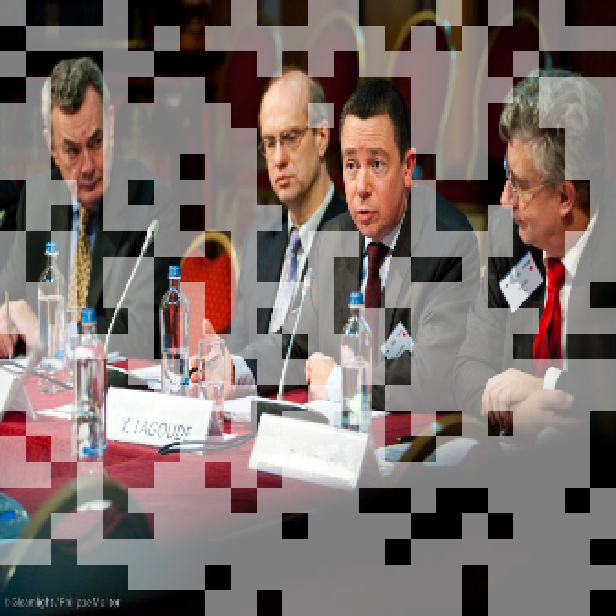}
        \caption{240}
    \end{subfigure}
    \hfill
    \begin{subfigure}[b]{0.3\textwidth}
        \centering
        \includegraphics[width=\textwidth]{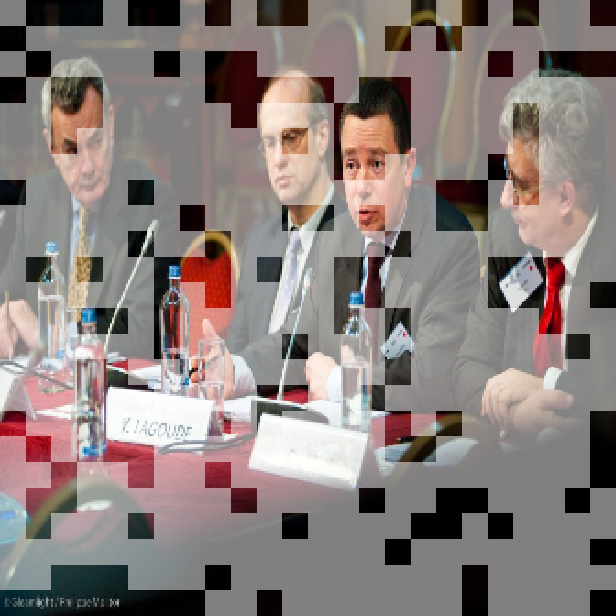}
        \caption{156}
    \end{subfigure}
    \hfill
    \begin{subfigure}[b]{0.3\textwidth}
        \centering
        \includegraphics[width=\textwidth]{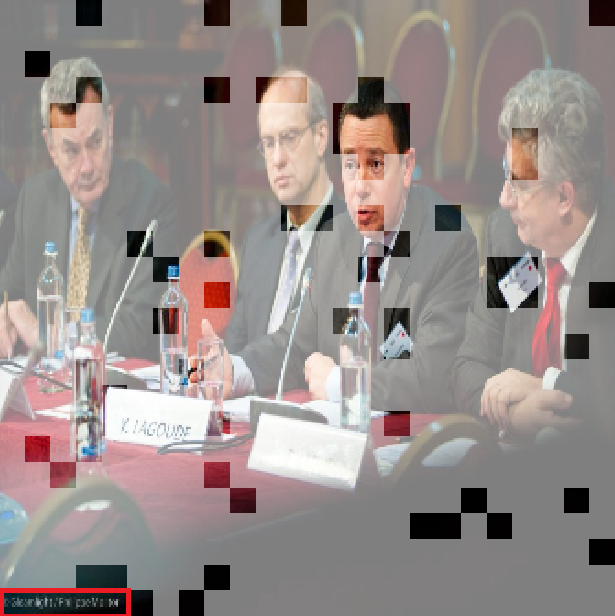}
        \caption{72}
    \end{subfigure}

    \caption{\textbf{Question}: Who was the photographer? {\color{blue}Ground\_truth}: Philippe molitor. {\color{green}Output}: Philippe molitor.(\Checkmark)}
    \label{fig7}
\end{figure*}

\begin{figure*}[h]
    \centering
    \begin{subfigure}[b]{0.3\textwidth}
        \centering
        \includegraphics[width=\textwidth]{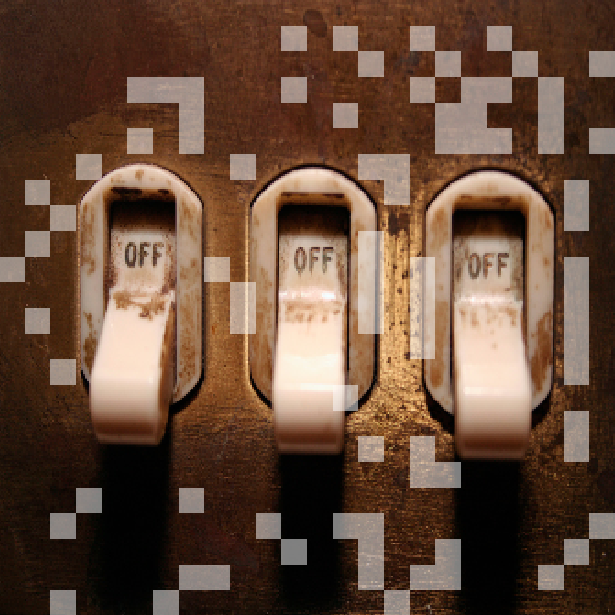}
        \caption{492}
    \end{subfigure}
    \hfill
    \begin{subfigure}[b]{0.3\textwidth}
        \centering
        \includegraphics[width=\textwidth]{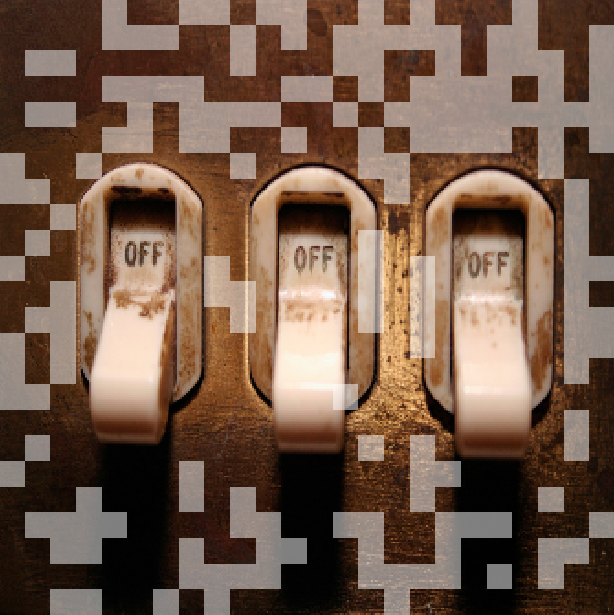}
        \caption{408}
    \end{subfigure}
    \hfill
    \begin{subfigure}[b]{0.3\textwidth}
        \centering
        \includegraphics[width=\textwidth]{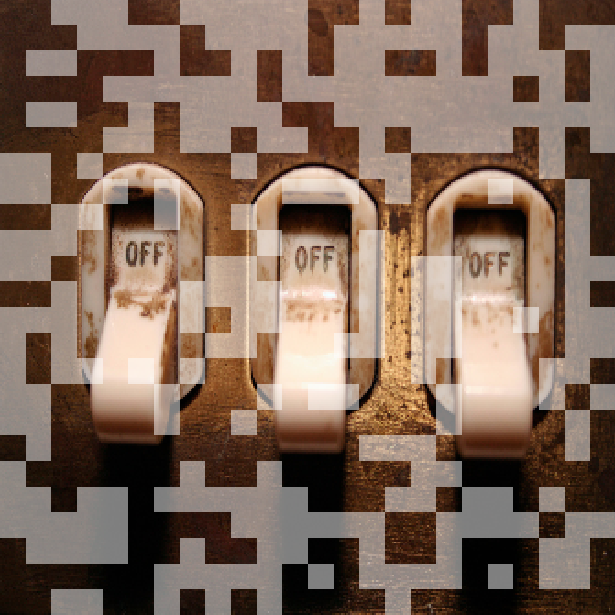}
        \caption{324}
    \end{subfigure}
    \begin{subfigure}[b]{0.3\textwidth}
        \centering
        \includegraphics[width=\textwidth]{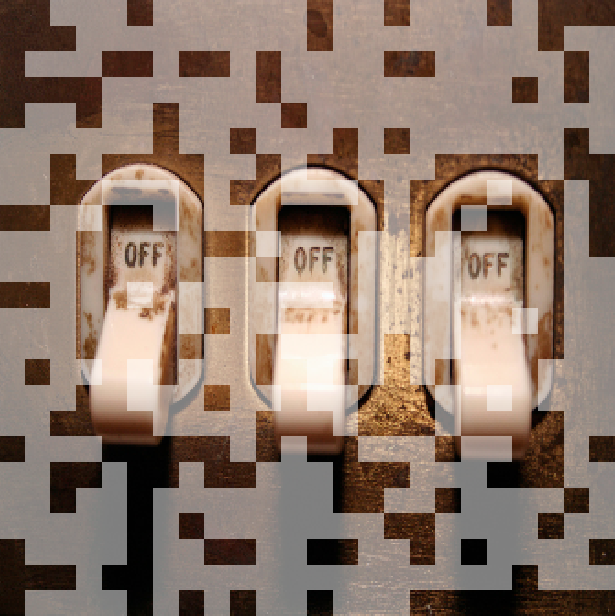}
        \caption{240}
    \end{subfigure}
    \hfill
    \begin{subfigure}[b]{0.3\textwidth}
        \centering
        \includegraphics[width=\textwidth]{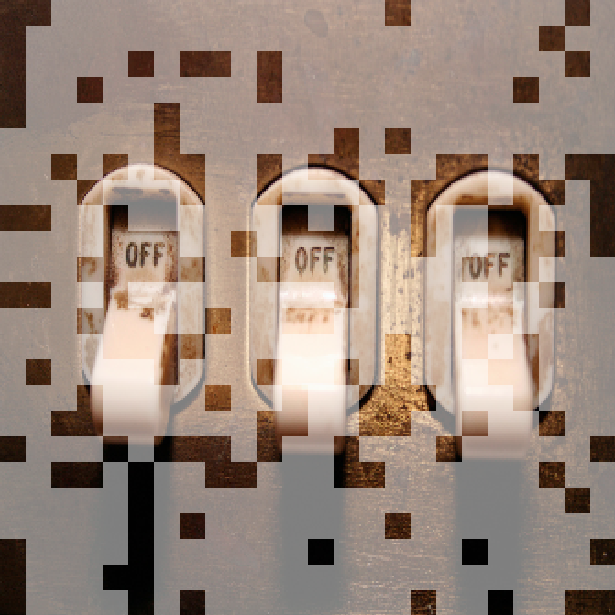}
        \caption{156}
    \end{subfigure}
    \hfill
    \begin{subfigure}[b]{0.3\textwidth}
        \centering
        \includegraphics[width=\textwidth]{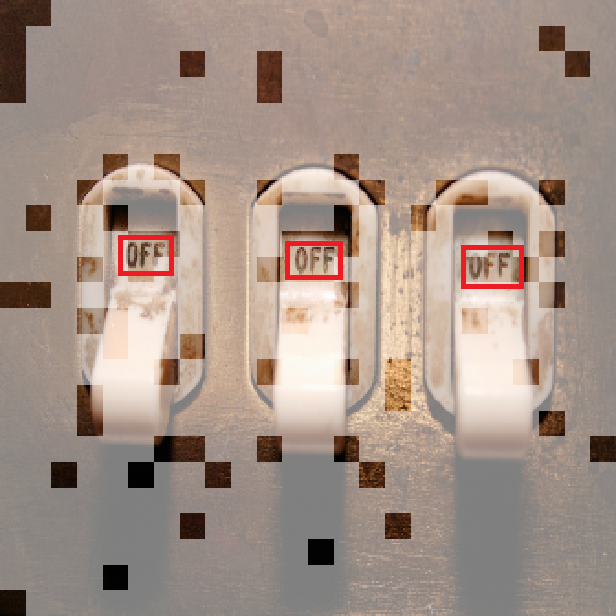}
        \caption{72}
    \end{subfigure}
    \caption{\textbf{Question}: Are these switches on or off? {\color{blue}Ground\_truth}: off. {\color{green}Output}: off.(\Checkmark)}
    \label{fig8}
\end{figure*}

\begin{figure*}
    \centering
    \begin{subfigure}[b]{0.3\textwidth}
        \centering
        \includegraphics[width=\textwidth]{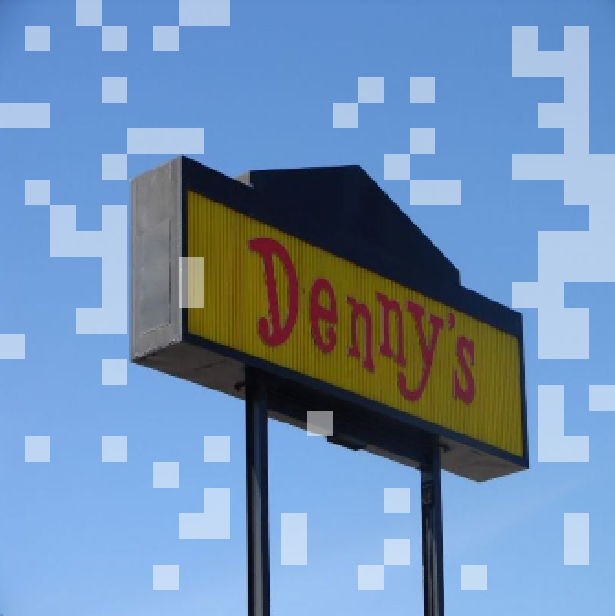}
        \caption{492}
    \end{subfigure}
    \hfill
    \begin{subfigure}[b]{0.3\textwidth}
        \centering
        \includegraphics[width=\textwidth]{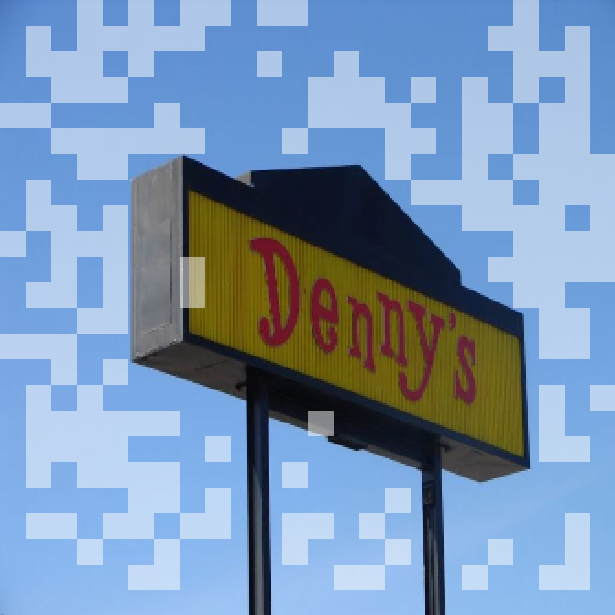}
        \caption{408}
    \end{subfigure}
    \hfill
    \begin{subfigure}[b]{0.3\textwidth}
        \centering
        \includegraphics[width=\textwidth]{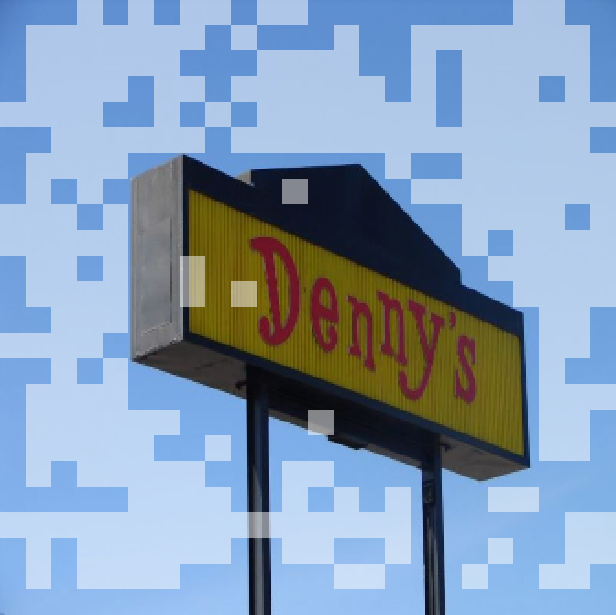}
        \caption{324}
    \end{subfigure}
    \begin{subfigure}[b]{0.3\textwidth}
        \centering
        \includegraphics[width=\textwidth]{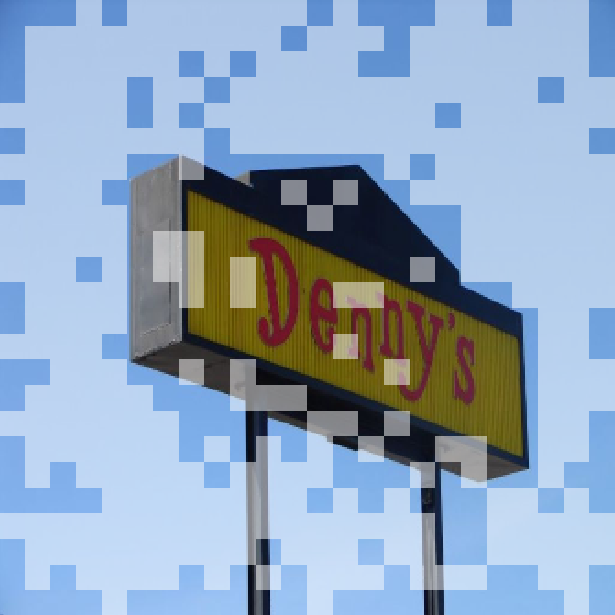}
        \caption{240}
    \end{subfigure}
    \hfill
    \begin{subfigure}[b]{0.3\textwidth}
        \centering
        \includegraphics[width=\textwidth]{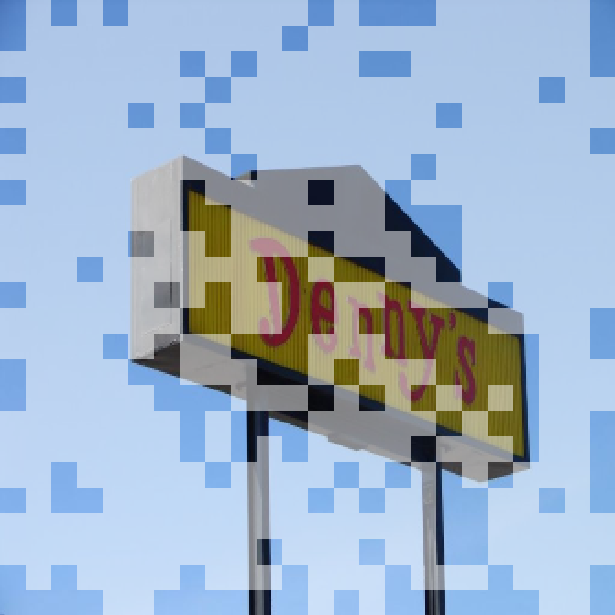}
        \caption{156}
    \end{subfigure}
    \hfill
    \begin{subfigure}[b]{0.3\textwidth}
        \centering
        \includegraphics[width=\textwidth]{./visual/5ce862cbefd8458f_color/72.png}
        \caption{72}
    \end{subfigure}
    \caption{\textbf{Question}: What color are the letters on this sign? {\color{blue}Ground\_truth}: Red. {\color{green}Output}: Red.(\Checkmark)}
    \label{fig9}
\end{figure*}
\end{document}